\documentclass{article}
\usepackage{spconf,amsmath,epsfig}
\usepackage[utf8]{inputenc} 
\usepackage[T1]{fontenc}    
\usepackage{hyperref}       
\usepackage{url}            
\usepackage{booktabs}       
\usepackage{amsfonts}       
\usepackage{amsmath}
\usepackage{amssymb}
\usepackage{nicefrac}       
\usepackage{microtype}      
\usepackage{xcolor}         

\usepackage{multirow}
\usepackage{adjustbox}

\usepackage{graphicx}
\usepackage{colortbl} 

\title{Bootstrapping Rare Object Detection in High-Resolution Satellite Imagery}

\name{\begin{tabular}{c}
Akram Zaytar\sthanks{Corresponding author: \texttt{akramzaytar@microsoft.com}}\textsuperscript{1}, Caleb Robinson\textsuperscript{1}, Gilles Q. Hacheme\textsuperscript{1}, Girmaw A. Tadesse\textsuperscript{1}, Rahul Dodhia\textsuperscript{1} \\
Juan M. Lavista Ferres\textsuperscript{1}, Lacey F. Hughey\textsuperscript{2}, Jared A. Stabach\textsuperscript{2}, Irene Amoke\textsuperscript{3}
\end{tabular}}

\address{Microsoft AI for Good Research Lab\textsuperscript{1}  \\ Smithsonian National Zoo \&
Conservation Biology Institute\textsuperscript{2} \\ Kenya Wildlife Trust\textsuperscript{3}}

\begin{document}
\maketitle

\begin{abstract}
Rare object detection is a fundamental task in applied geospatial machine learning, however is often challenging due to large amounts of high-resolution satellite or aerial imagery and few or no labeled positive samples to start with. This paper addresses the problem of bootstrapping such a rare object detection task assuming there is no labeled data and no spatial prior over the area of interest. We propose novel offline and online cluster-based approaches for sampling patches that are significantly more efficient, in terms of exposing positive samples to a human annotator, than random sampling. We apply our methods for identifying bomas, or small enclosures for herd animals, in the Serengeti Mara region of Kenya and Tanzania. We demonstrate a significant enhancement in detection efficiency, achieving a positive sampling rate increase from 2\% (random) to 30\%. This advancement enables effective machine learning mapping even with minimal labeling budgets, exemplified by an $F_{1}$ score on the boma detection task of $0.51$ with a budget of $300$ total patches.
\end{abstract}

\begin{keywords}
Geospatial machine learning, Rare Object Detection, Semantic Segmentation, Remote Sensing, Boma Detection.
\end{keywords}

\section{Introduction}

Rare object detection over remotely sensed (satellite, aerial, drone) imagery is a common task in geospatial machine learning with applications ranging from identifying damaged structures in post-disaster imagery~\cite{joshi2017damage, robinson2023rapid}, renewable energy infrastructure mapping at country and global scales~\cite{hu2021synthetic}, to ecological studies like counting large mammals over vast landscapes~\cite{wu2023deep}. The nature of such tasks is that they are often not accompanied by relevant labeled datasets -- if such datasets existed over the area of interest (AOI), then there would not be a need to solve the task in the first place. As such, a first step in any rare object detection task is often manually finding \textit{few} instances of the rare object, i.e.,  positive class examples, that can then be used to \textit{bootstrap} a few-shot model-based approach~\cite{goupilleau2021active,zhang2024few}.

It is possible to incorporate known spatial priors into the \textit{bootstrapping} process in some object detection tasks. For example, to find examples of damaged structures in post-disaster satellite imagery, it is possible to search over known existing structures instead of over the entire AOI, dramatically reducing the search space. However, in other problem instances, such as finding large mammals in high-resolution satellite/aerial imagery, the expected spatial distribution of the object of interest is less obvious.

Similarly, in some problem instances, it is possible to substitute related label datasets to bootstrap the modeling process. For example, OpenStreetMap (OSM) contains a large amount of data on solar photovoltaic plants and windmills~\cite{dunnett2020harmonised} that can be joined with satellite imagery to train machine learning models that can then identify solar panels and windmills in imagery. However, such models will be limited by the coverage of existing labels and accompanying imagery. For example, the coverage of OSM data varies greatly by country, and imagery from different countries can vary greatly, therefore models trained under such conditions may easily fail to generalize over the entire AOI. Domain adaptation methods using synthetic imagery~\cite{hu2021synthetic, martinson2021training}, strong augmentation~\cite{zhao2021improved}, and near-class object detectors~\cite{lee2015fine} have all been proposed to alleviate this problem, but the fundamental problem of how to \textit{start} modeling given a novel application and no labels remains an open problem in the state-of-the-art rare object detection literature.

In this paper, we formulate and address the basic problem of \textit{bootstrapping a dataset of positive rare object samples under the assumptions of no initial labeled data and no spatial priors}. We propose novel offline and online clustering-based methods for selecting initial patches to annotate that only depends on imagery inputs, and a way to set the parameters of these methods without labels. Generally, our approach relies on the intuition that rare objects will, by definition, appear differently than their surroundings.

We apply our approach on the real-world problem of identifying Bomas from satellite imagery in the Serengeti Mara, a region of high ecological importance in Kenya and Tanzania. The distribution of Bomas is critical for various ecological and conservation efforts~\cite{tyrrell2022landscape}. Our approach significantly increased the positive sampling rate from 2\% to 30\%, this sampling efficiency translates to the downstream task of Boma detection, achieving an $F_{1}$ score of $0.51$ with just $300$ initial labels.

The implications of this research extend beyond the specific case study, offering a scalable and efficient framework for rare object detection in various geospatial and remote sensing applications.

\section{Problem Formulation}\label{sec:problem_statement}

We assume that we are given a large unlabeled high-resolution satellite imagery scene, $\mathbf{X}$, that covers some AOI and a limited labeling budget, $b$ (i.e., total area of imagery that can be labeled). We would like to detect some instances of a rare object class in $\mathbf{X}$ in order to \textit{bootstrap} a modeling process.
Specifically, we are looking for $n^+$ \textit{positive} instances, i.e. examples of the object. While we look for the positives, we will annotate a number of \textit{negative} instances, $n^-$, i.e. where the rare object is not present. Our objective is thus to minimize $\frac{n^-}{n^+}$ using $b$ labelling iterations (or budget).

To achieve this, we split $\mathbf{X}$ into an $H \times W$ grid of non-overlapping image patches. For each grid cell, $X_{i,j}$, we assign a probability, $P_{i,j}$, thus initializing a \textit{sampling surface}, $\mathcal{P}$ (i.e. a discrete probability distribution on a 2D grid). We discuss different strategies for initializing $\mathcal{P}$ in Section \ref{sec:methods}. We then use a sampling strategy to choose $b$ patches for a labeler to annotate.

\section{Methods}\label{sec:methods}

\begin{figure*}
    \centering
    \includegraphics[width=0.80\textwidth]{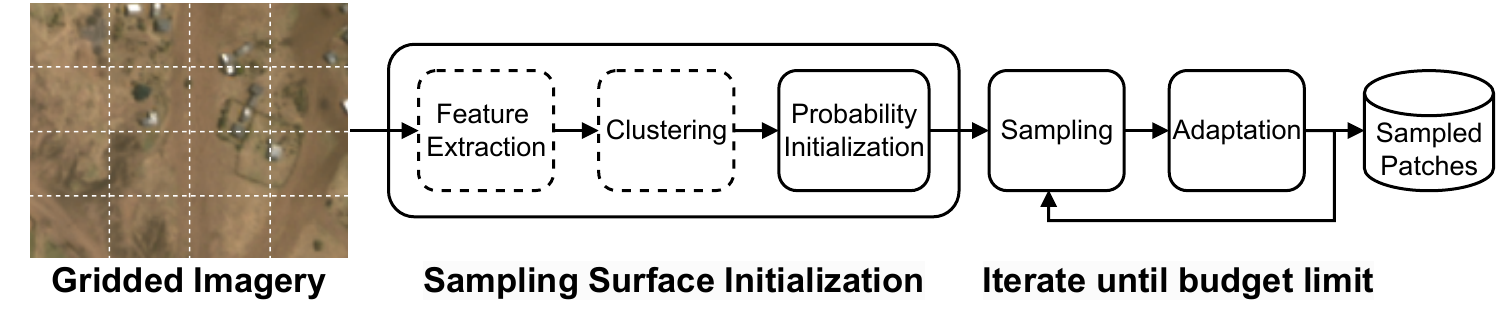}
    \caption{\textbf{Bootstrapping rare object detection}. Given input imagery, we create a grid of image patches, initialize a sampling surface over the same grid, and sample iteratively from the surface looking for rare object instances until we hit a budget limit. Sampling strategies that surface these rare objects more frequently than random allow for quicker instantiating of model based methods for finding such objects.}
    \label{fig:bdg}
\end{figure*}

Our framework for bootstrapping rare object detection in $\bf{X}$ depends on two steps: \textit{initializing the sampling surface}, and defining a \textit{sampling strategy} to select $b$ patches while, optionally, updating the sampling surface (as shown in Fig.~\ref{fig:bdg}).

\subsection{Initializing the sampling surface}
Naively, we can use equal weights to initialize the sampling surface as a \texttt{Uniform} baseline, i.e., $P_{i,j}=\frac{1}{H*W}, \forall i,j$. We further propose cluster-based approaches that extract feature vectors for each image patch $X_{i,j}$ using 3 different strategies: 1) \texttt{RCF}---uses random convolutional features~\cite{rolf2021generalizable} to extract color/texture feature vectors; 2) \texttt{ColorStats}---calculates the mean, standard deviation, minimum, and maximum values for each channel in $X_{i,j}$, providing a simple representation of the colors in a patch; 3) \texttt{Pre-trained ResNet}---employs a pre-trained ResNet-18~\cite{he2016deep} to extract a feature representation. The clustering step assigns each grid cell into one of several clusters based on its feature representation. Here we utilize \texttt{KMeans} \& \texttt{Bisecting K-Means}~\cite{steinbach2000comparison} with a hyperparameter for the number of clusters, $K$, and \texttt{DBSCAN}~\cite{ester1996density} with hyperparameters for the maximum distance between two samples to be neighbors, $\epsilon$, and the number of neighbors of a point to be considered a core point, $\eta$. We describe an unsupervised method for choosing these hyperparameters based on the silhouette score~\cite{rousseeuw1987silhouettes} in Appendix Section 1.

We use the feature representations per patch to fit the clustering algorithm, resulting in $K$ clusters, where each $X_{i,j}$ is assigned to one of the clusters. We then initialize $P_{i,j}$ based on the \textit{inverse size of the cluster that $X_{i,j}$ is in}. Specifically, let $C_{i,j}$ be the size of the cluster that $X_{i,j}$ is in, then:
\begin{equation}
P_{i, j} \leftarrow \frac{1}{K} * \frac{1}{C_{i,j}}, \forall i,j
\end{equation}

\subsection{Sampling strategies}
Once $\mathcal{P}$ is initialized, we can sample from it with methods from two broad categories: 1.) offline sampling where $P_{i,j}$ do not change based on incoming online annotations; and 2.) online sampling where we use the incoming annotations to change the probability surface. Specifically, we propose the \texttt{Online} and \texttt{Proximity} methods. In the \texttt{Proximity} sampling method, if we sample a positive, then we increase the probability of all neighboring patches within a set radius, $r$, by some weight, $w$, following the intuition that rare objects may be clustered in space. Similarly, in the \texttt{Online} method we increase the probability of patches that are in the same cluster as the sampled positive by $w$. In both cases, we sample from $\mathcal{P}$ without replacement and renormalize after reweighting based on observed positives. 

With both online and offline methods, we sample until we exhaust our budget, after which we can use the set of found positives, negatives, and unlabeled patches to train a downstream machine learning model for object detection.

\section{Case Study: Boma Mapping in the Serengeti Mara}\label{sec:sim}
We apply our methods in a case study for finding ``bomas'' in high-resolution satellite imagery captured over the Serengeti Mara region of Kenya and Tanzania. Bomas are temporary cattle enclosures used by local population that are relatively rare given the low population density of the region. Information on boma locations are crucial for identifying human-predator conflict hotspots and, as such, are used by organizations such as the Kenya Wildlife Trust.

We use 3 pansharpened WorldView-2 satellite scenes with $50\text{cm}/\text{px}$ spatial resolution. Combined, the images cover approximately $4,300$ $km^2$ over three points in time (August 6th, 2022, January 2 \& October 8 2020). We have polygon based labels for all bomas in these scenes from prior work which we use to run simulations.

\subsection{Bootstrapping}
For our experiments we choose $3$ \emph{low-resource} labeling budgets of $300$, $950$, and $3000$ image patches to evaluate different sampling strategies. For \texttt{Proximity} weighting, the radius was set at $r=200$ $m$, informed by previous knowledge of Boma settlement patterns.
For both \texttt{Proximity} and \texttt{Online}, a value of $w = \max(P_{ij})$ (highest initial weight) was used.
For clustering-based methods for initializing the sampling surface, we set hyperparameters (including which feature representation to use) based on an \textit{unsupervised} method described in Appendix Section 1 that attempts to create clusters of rare objects. After setting the hyperparameter values, we compared the performance of various sampling approaches (i.e., \texttt{Uniform}, \texttt{Proximity}, and clustering methods in both \emph{Offline} and \emph{Online} scenarios using three labeling simulations per method. We report the number of positives samples found with each method in Table \ref{tab:results}. 

\subsection{Downstream training}
Each bootstrapping method produces a set of positive and negative labels that were used to create training sets for the downstream task of semantic segmentation of bomas. Our aim is to assess the benefit provided by each method in the task of rare object detection. To this end, we report object-level $F_{1}$ scores over unseen patches in Table \ref{tab:results}. We maintained a consistent training setup across all methods, utilizing a U-Net~\cite{ronneberger2015u} architecture with a ResNeXt50 ($32x4d$) backbone~\cite{xie2017aggregated}, and training for up to 200 epochs. The AdamW optimizer was employed alongside data augmentation techniques, including flipping, rotation, and color jitter. Our training approach encompassed two loss configurations: 1) traditional cross-entropy loss (\texttt{CE}), and 2) regularized cross entropy (RCE) loss - a hybrid loss that combines cross-entropy for labeled pixels with an entropy minimization regularization term for unlabeled patches, formulated as $J(y,\hat{y})=\rho(CE(y,\hat{y}) \odot Y_{L}) + (1 - \rho)(H(\hat{y}) \odot Y_{\bar{L}})$. Here, $\rho$ represents the proportion of labeled pixels, $H$ signifies entropy, $Y_{L}$ is the binary mask for labeled pixels, and $Y_{\bar{L}}$ indicates the mask for unlabeled pixels. This is a semi-supervised training technique based on~\cite{grandvalet2006entropy} that aims to minimize the class entropy of predictions over unlabeled pixels, which stabilizes training in low-label settings.

\begin{table*}[]
\centering
\resizebox{\textwidth}{!}{%
\begin{tabular}{@{}lccccccccc@{}}
\toprule
\multicolumn{1}{c}{\multirow{2}{*}{\textbf{Sampling Strategy}}} & \multicolumn{3}{c}{\textbf{300 Patches}} & \multicolumn{3}{c}{\textbf{950 Patches}} & \multicolumn{3}{c}{\textbf{3K Patches}} \\ \cmidrule(lr){2-4} \cmidrule(lr){5-7} \cmidrule(lr){8-10}
\multicolumn{1}{c}{} & $\mathbf{n^+}$ & \textbf{CE} & \textbf{RCE} & $\mathbf{n^+}$ & \textbf{CE} & \textbf{RCE} & $\mathbf{n^+}$ & \textbf{CE} & \textbf{RCE} \\ \midrule
Uniform & \;\;5 ± 0.8 & .01 ± .00 & .15 ± .07 & 15 ± 3 & .04 ± .16 & .57 ± .07 & 56 ± 9.8 & .59 ± .00 & .67 ± .00 \\
Proximity Weighting & 42 ± 8.6 & .00 ± .00 & .07 ± .03 & 157 ± 61 & .09 ± .16 & .48 ± .20 & 422 ± 34 & \bf{.73 ± .00} & .76 ± .00 \\ \midrule
KMeans & 40 ± 10 & .01 ± .00 & .02 ± .01 & 275 ± 78 & .22 ± .15 & .68 ± .01 & 883 ± 127 & .50 ± .19 & .76 ± .02 \\
Online KMeans & \textbf{85 ± 26} & \bf{.12 ± .00} & .35 ± .00 & 310 ± 73 & .13 ± .00 & .50 ± .10 & 845 ± 68 & .48 ± .20 & .75 ± .02 \\ \midrule
DBSCAN & 10 ± 2.3 & .00 ± .00 & .01 ± .00 & 27 ± 2 & .00 ± .00 & .37 ± .08 & 120 ± 2.35 & .70 ± .00 & .69 ± .00 \\
Online DBSCAN & 10 ± 1.4 & .00 ± .00 & .03 ± .00 & 50 ± 11 & .07 ± .00 & .33 ± .15 & 215 ± 7.5 & .10 ± .00 & .60 ± .00 \\ \midrule
Bisecting KMeans & 18 ± 2.6 & .03 ± .16 & .39 ± .16 & 58 ± 2.5 & .18 ± .08 & .71 ± .03 & 169 ± 5.7 & .72 ± .03 & .75 ± .01 \\
Online BKMeans & 57 ± 9.7 & .01 ± .00 & \textbf{.51 ± .16} & \textbf{371 ± 82} & \bf{.34 ± .22} & \textbf{.72 ± .02} & \textbf{1008 ± 145} & .60 ± .10 & \textbf{.78 ± .00} \\ \bottomrule
\end{tabular}%
}
\caption{Results of various sampling strategies across different labeling budgets (300, 950, and 3,000 patches). We report the number of positive instances identified ($\mathbf{n^+}$) by the sampling strategy and the downstream task performance of a U-Net model trained with the resulting samples. We report $F_1$ scores at an object level for models trained with Cross Entropy (CE) and Regularized Cross Entropy (RCE) losses. Note BKMeans refers to Bisecting KMeans.}
\label{tab:results}
\end{table*}

\subsection{Results and Discussion}\label{subsec:results}
Our experiments demonstrate that all methods significantly surpass the uniform sampling baseline in identifying rare objects (see Table~\ref{tab:results}). Notably, \texttt{Online Bisecting KMeans} increased the sampling ratio from 2\% (i.e., the object's density over the AOI) to $40\%$ in the 950 patch scenario and $30\%$ in the 3,000 patch scenario. In the 950 patch scenario, the \texttt{Online Bisecting KMeans} method found $25\times$ as many positives as uniform sampling---human annotators would need to process an additional 23,750 patches under the uniform sampling method to find an equivalent number of positives.

For the downstream task of detecting Bomas, the improved sampling strategies translated into notable gains in object detection performance, especially at lower labeling budgets. For instance, while uniform sampling (even with proximity weighting) fails with a budget of 300 patches, Online Bisecting KMeans achieves an $F_1$ score of $0.51$. Overall, the cluster-based and online sampling methods achieve the best down-stream performance with a best result of $0.78$ $F_1$ at the largest labeling budget. We find the advantage of these sampling techniques diminish as more labels become available, which is where other search methods can also take over.

Finally, we find that the inclusion of the entropy regularization term (RCE) consistently boosted performance across all experiments. While not the focus of this paper, this result warrants further exploration across other low-label modeling problems.

\section{Conclusion}
In this paper, we formalized the challenge of label bootstrapping for rare object detection in satellite imagery, establishing benchmark results for both heuristic-based methods, like proximity weighting, and clustering-based approaches. In our case study, the proposed methods significantly improved the positive sample identification rate from $2\%$ to $30\%$, while enabling machine learning mapping with minimal labeling resources, achieving an F1 score of $0.51$ with just $300$ labeled patches. We hope this work paves the way for the exploration of bootstrapping strategies for geospatial ML and their connections to adjacent techniques in active learning and subset selection.

\bibliographystyle{IEEEbib}
\bibliography{strings,main}

\clearpage
\appendix

\section{Unsupervised method for choosing hyperparameters}

Our proposed methods for initializing a sampling surface involve 1.) computing some feature representation per image patch (we test three methods); and 2.) clustering these representations in some manner (a method that will require some hyperparameter choice, e.g. $K$ in KMeans). As we assume that we have no access to labeled data, or more generally, any prior information about the distribution of the object of interest, there is a large question of how to set the parameters of our method. Here we propose an \textit{unsupervised} approach based on the silhouette score~\cite{rousseeuw1987silhouettes}.

\cite{rousseeuw1987silhouettes} define a \textit{silhouette coefficient} given a clustering of data and a data point $x_i$ that is the ratio $s_i = \frac{b_i-a_i}{\max\left(a_i,b_i\right)}$. Here $a_i$ is the mean intra-cluster distance of $x_i$, i.e. the average distance to data points in its own cluster, and $b_i$ is the mean nearest cluster distance, i.e. the average distance to all data points in the nearest cluster. The \textit{silhouette score} is the average silhouette coefficient over all data points in a dataset and can range from $-1$ (indicating that all samples are in wrong clusters, i.e. closer to other clusters than to the cluster they are assigned) to $1$ (indicating that samples are well clustered).

\begin{figure}[ht]
    \centering
    \includegraphics[width=0.5\textwidth]{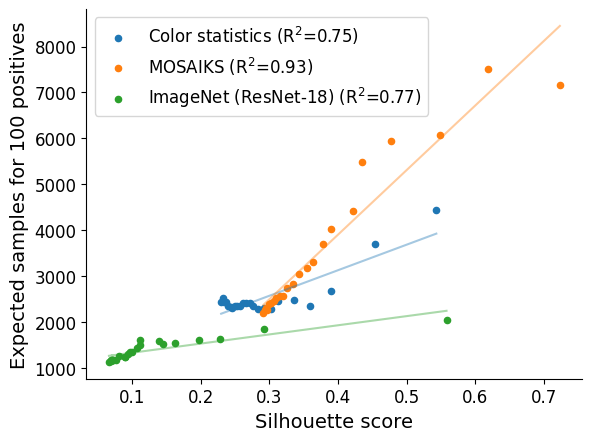}
    \caption{Silhouette scores of different clustering methods versus the expected number of samples required to find 100 positives with an \texttt{offline} sampling scheme.}
    \label{fig:silhouette}
\end{figure}

We find that the silhouette score from a choice of feature representation and clustering method hyperparameters is highly correlated with the number of samples required to find 100 positive samples. For example in Figure \ref{fig:silhouette} we show the number of samples required to find 100 positive samples with the \texttt{Offline KMeans} for different feature representations and choices for $k$. Overall, the expected number of samples is minimized when the silhouette score is also minimized, and there is high correlation between the two for each feature representation  (e.g. $R^2 = 0.93$ when using MOSAIKS based representations). Given this, we use a Bayesian hyperparameter search to minimize the absolute value of the silhouette score given all free parameters, and use the resulting values in each experiment.

\end{document}